%% file: main.tex
\documentclass{article}

\usepackage{algorithmic}
\usepackage{algorithm}

\PassOptionsToPackage{numbers, compress}{natbib}
\usepackage[preprint]{neurips_2023}

\usepackage[utf8]{inputenc} % allow utf-8 input
\usepackage[T1]{fontenc}    % use 8-bit T1 fonts
\usepackage{url}            % simple URL typesetting
\usepackage{booktabs}       % professional-quality tables
\usepackage{amsfonts}       % blackboard math symbols
\usepackage{nicefrac}       % compact symbols for 1/2, etc.
\usepackage{microtype}      % microtypography
\usepackage{xcolor}         % colors

\usepackage{amsmath}
\usepackage{amssymb}
\usepackage{mathtools}
\usepackage{amsthm}
\usepackage{xcolor} 
\usepackage{graphicx}
\usepackage{caption}
\usepackage{subcaption}
\usepackage{sgame}
\usepackage{color}
\usepackage{url}
\usepackage{setspace}
\usepackage{enumitem}
\usepackage{etoolbox}
\usepackage{comment}
\theoremstyle{definition}

\usepackage{hyperref}       % hyperlinks

\title{Uncertainty Quantification for Local Model Explanations Without Model Access}
\author{%
  Surin Ahn\thanks{Work completed during an internship at Microsoft.}\\
  Stanford University\\
  \texttt{surinahn@stanford.edu} \\
\And
  Justin Grana\thanks{Work completed while a full-time employee at Microsoft.}\\
  Edge \& Node\\
  \texttt{justin@edgeandnode.com} \\
\And
Yafet Tamene\\
Microsoft\\
\texttt{yafettamene@microsoft.com}\\
\And
Kristian Holsheimer\footnotemark[2]\\%\thanks{Work completed while a full-time employee at Microsoft.}\\
Google DeepMind\\
\texttt{holsheimer@google.com}
}%

\begin{document}

\maketitle

\begin{abstract}
  %   We present a bootstrap algorithm for estimating local interpretable model-agnostic explanations for a machine learning model without direct access to the model but instead only a sample of model inputs and outputs.
  We present a model-agnostic algorithm for generating post-hoc
  explanations and uncertainty intervals for a machine learning model
  when only a static sample of inputs and outputs from the model is
  available, rather than direct access to the model itself.
  %direct access to the model is unavailable, and instead only a fixed sample of inputs and outputs from the model is given to the explainer. 
  This situation may arise when model evaluations are expensive; when
  privacy, security and bandwidth constraints are imposed; or when
  there is a need for real-time, on-device explanations. Our algorithm uses a bootstrapping approach to quantify the uncertainty that inevitably arises when generating explanations from a finite sample of model queries. 
  %Our algorithm constructs explanations using local regression and quantifies the uncertainty of the explanations using a bootstrapping approach.  We focus on uncertainty quantification and
  %Our approach is to use a bootstrap in which we estimate both the explanation as well as quantify the uncertainty associated with those explanations.  
  Through a simulation study, we show that the uncertainty intervals
  generated by our algorithm exhibit a favorable trade-off between
  interval width and coverage probability compared to the naive
  confidence intervals from classical regression analysis as well as
  current Bayesian approaches for quantifying explanation uncertainty.
  %Using simulated data, we show that our bootstrap algorithm performs better than the frequentist theoretical confidence intervals that have insufficient coverage.
  We further demonstrate the capabilities of our method by applying it
  to black-box models, including a deep neural network, trained on three real-world datasets.\footnote{GitHub repository: \url{https://github.com/surinahn/xai-uncertainty-quantification}} 
\end{abstract}

\input{intro_lit}
\input{setup}
\input{model}

\input{results}

\input{realdata}

\input{future}

\bibliographystyle{apalike}
\bibliography{refs.bib}

%\iffalse
\newpage
\input{supp}
%\fi

\end{document}

%% file: intro_lit.tex
%\section*{Problem Statement and Background}
%\vspace{-1.5cm}
%\section{Introduction, Problem Statement and Related Work}
\section{Introduction} 
The advent and deployment of %production-grade 
machine learning (ML) and
artificial intelligence (AI) systems across numerous sectors have elevated the need to understand
and explain such systems.  Transparency and explainability of ML models are critical for debugging 
their mistakes, investigating their bias and fairness \cite{predictive1}, building user trust in their decision-making, 
and in general making them more ``human-centric'' \cite{riedl2019human}. 
For example, explainability is crucial
when determining why a malware classifier
missed malicious content \cite{malware}, examining errors in the
behavior of an autonomous vehicle \cite{auton1}, or conveying to a person why their 
loan application was denied \cite{loan1,loan2}. 
%or investigating model bias and fairness \cite{predictive1}. 
The profound societal implications of deploying
``black-box'' systems have spurred regulations such as the European
Union's General Data Protection Regulation (GDPR), which purportedly
establishes explanations of algorithmic decision making as a
fundamental human right \cite{goodman2017european}.
Recently, the White House released a blueprint for an ``AI Bill of Rights'', 
outlining a plan for enforcing the accountability, explainability and trustworthiness of AI systems \cite{aibill}. 
%While developing model understanding is
%a technical challenge, the social implications of deploying ``black-box'' systems are increasingly topical in the policy sphere, punctuated by the ``Right to an Explanation \cite{goodman2017european},'' codified into law in the European Union.

The demand for model explainability has given rise to several methods that provide practitioners with insights into the key drivers of an ML model.
%While some approaches suggest outright abandoning black-box models and instead fitting flexible yet human understandable models \cite{ebm},
While some methods take the approach of replacing black-box models with flexible, human-interpretable models \cite{ebm},
others such as LIME \cite{lime1}, SHAP
\cite{lundberg2017unified}, and ROAR \cite{roar} focus on constructing post-hoc explanations of black-box models.  
Given the
%practical and %policy
technical and societal importance of model explainability, it is no surprise that
additional remedies continue to arise (see surveys
\cite{surv1,surv2,surv3,surv4,surv5}).  Furthermore, the potential applications of model explainability are expanding rapidly, from sports predictions \cite{baseball} to cancer diagnosis \cite{cancer}.

A major limitation of many post-hoc explainers is that they require direct
access to the underlying model, i.e., the ability to query the model
arbitrarily, in real-time. % For example,
% LIME and SHAP estimate the importance of each feature on the
% prediction by observing the model outputs under numerous local
% perturbations of the input instance. 
However, there are numerous situations where model access is restricted
or entirely unavailable. Some inhibiting factors---all of which are
especially prevalent in production-grade systems---include the
complexity of the engineering pipeline, model privacy and security
concerns, and the need for real time, on-device explanations. Instead,
it is often more feasible to collect a batch of input-output samples
from the model in advance and employ this static dataset to construct
explanations.  In this environment, the fidelity of the explanation is
limited by the available data, and thus it is critical to quantify the uncertainty in the explanation. \emph{Motivated by this problem, we (i) show how to apply local model-agnostic explainers on a
  fixed dataset of model inputs and outputs, without the need for
  direct model access; (ii) quantify the uncertainty associated with the explanations using a non-parametric bootstrapping approach; (iii)
  %demonstrate how our method out-performs current benchmarks in a simulation study; 
  demonstrate the favorable performance of our method relative to current benchmarks through a simulation study; 
  and (iv) apply our method to
  several real-world datasets and models, including an attention-based deep neural network.} 
  
  Our primary contribution is our bootstrap-based uncertainty quantification technique, which can be used with any local explainer in a ``plug-and-play'' fashion. 
  %While adapting local model-agnostic explanations to a fixed dataset is necessary, the core of our contribution is in quantifying the uncertainty associated with the explanations. 
  Our comparison with a (Bayesian) parametric method \cite{reliable1} for computing explanation uncertainty is especially salient. Since local surrogate models are usually misspecified, error in the explanation stems from model misspecification in addition to standard sampling error. We show that our non-parametric bootstrap method is better at accounting for this type of error than standard parametric approaches that assume the model is correctly specified up to some mean-zero noise.

\paragraph{Other Related Work} Our approach builds upon previous
methods that attempt to quantify the uncertainty of explanations.
These include methods that apply standard sampling error techniques to
explainability \cite{reliable1,howmuch} as well as non-parametric
approaches \cite{ordinal,clinical}.  While \cite{ordinal} also employs bootstrap methods, we focus on
gradient estimation (as in \cite{ucam}) instead of rank
orders of feature importance. Additionally, \cite{clinical} focuses on
quantifying the uncertainty in explanations of convolutional neural
networks, while our method is model-agnostic.  CXPlain \cite{cxplain}
also proposes to use the bootstrap, but this work does not use the same definition of an explanation that we do, consider an environment without model access, systematically compare the bootstrap with other approaches, or provide guidance on setting the bootstrap hyperparameters.
%using the bootstrap, but that work is different in that they use a different definition of explanation, do not systematically compare the bootstrap with other approaches, do not give guiadance on how to set bootstrap hyper-parameters, and do not consider an environment without model access.  
To the best of our knowledge, the only other works that do not assume model access are \cite{art1,art2}, though they focus on Shapley values.

Finally, our work is tangentially related to the stability of ML model
explanations \cite{robust1,robust2,robust3}.  This research thrust
examines the sensitivity of explanations to perturbations of the input
data. Although in the environment we consider it is impossible to
observe the model outputs under arbitrary input perturbations, our
bootstrap resampling method still allows us to quantify the
sensitivity of explanations from a fixed dataset. It does so by
aggregating explanations computed from several local subsets of the
data (i.e., input samples generated from the empirical distribution of
the fixed dataset).  %\jg{Add all the other citations}

%% file: setup.tex
\section{Methodology}
% \jg{I don't see any reason why we couldn't do the same for vector valued ML models (think multi-class prediction) and just use the Jacobian. Maybe worth a footnote?  I don't think we need to include it in the formal writeup.}
 \paragraph{Setup} Let $f: \mathbb{R}^d \to \mathbb{R}$ denote a black-box ML model %, and let $\phi \triangleq \begin{bmatrix}\phi_1 & \ldots & \phi_d\end{bmatrix}^\top$ denote the vector of input features (which can be continuous or categorical). 
which maps a $d$-dimensional feature vector %(representing a set of continuous or categorical features) 
to some prediction.\footnote{Though we focus on models that produce a scalar output, our method can be straightforwardly extended to the case of vector outputs (e.g., a softmax layer for multiclass classification). } 
In classification tasks, $f(x) \in [0,1]$ represents the probability that $x$ belongs to a particular class. Instead of direct access to $f$, we are given a fixed dataset of $n$ input-output pairs from the model: $\mathcal{D} \triangleq \{(x^{(i)}, f(x^{(i)}))\}_{i=1}^n$.
Let $x^* \in \mathbb{R}^d$ be the input instance whose model prediction, $f(x^*)$, we would like to explain, and let $\pi_{x^*}: \mathbb{R}^d \to \mathbb{R}_+$ be a \textit{proximity function} which provides a measure of closeness to $x^*$. For example, LIME \cite{lime1} uses an exponential kernel applied to the $\ell_2$ distance. Given $\mathcal{D}$, our objective is two-fold.

\paragraph{Feature Importance Scores}  First, we would like to estimate the \textit{feature importance scores}, i.e., the influence of each feature on the model prediction. For simplicity and concreteness, and to ensure the availability of a ``ground truth'' in our experiments, we focus on estimating \textit{gradients} and \textit{local function differences}, which are commonly used proxies for feature importance \cite{baehrens2010explain, simonyan2013deep, springenberg2014striving, smilkov2017smoothgrad, ancona2019gradient}. The gradient of $f$ at $x^*$ %with respect to the feature vector $\phi$ 
is denoted by $\nabla_x f(x^*)$ and defined as: 
    \[\nabla_x f(x^*) \triangleq 
    \begin{bmatrix}
    \frac{\partial f}{\partial x_1}(x^*) &
    \frac{\partial f}{\partial x_2}(x^*) & 
    \cdots &
    \frac{\partial f}{\partial x_d}(x^*)
    \end{bmatrix}^\top.
    \]
Intuitively, if $\big\vert\frac{\partial f}{\partial x_j}(x^*)\big\vert$ is large, then a small change in the $j^\text{th}$ feature results in a large change in the model output, suggesting that the feature played a non-trivial role in the prediction. Note that when $f$ is a linear model, i.e., $f(x) = \beta_0 + \beta^\top x$ for some $\beta \in \mathbb{R}^d$, the gradient is simply the vector of coefficients $\beta$.

In certain situations (e.g., when features are categorical or ordinal; the model is non-differentiable; or the gradient is not a suitable measure of feature importance\footnote{Consider an ML model that exhibits step function-like behavior. Such models are not amenable to gradient-based explainability since the instantaneous derivatives are zero almost everywhere.}), it is useful to estimate the local function differences instead of the instantaneous partial derivatives. If the $j^\text{th}$ feature is continuous or ordinal, we define the function difference around $x^*$ with respect to the $j^\text{th}$ feature to be $f(x^*_+) - f(x^*_-)$,
%\[f(x^*_+) - f(x^*_-),\] 
where $x^*_+$ (resp. $x^*_-$) is equal to $x^*$ with the $j^\text{th}$ entry set to $x^*_j+\delta$ (resp. $x^*_j-\delta$), and $\delta$ is a domain-specific parameter specified by the user or set to some default constant that captures a ``one-unit change'' in that feature (e.g., a value proportional to the standard deviation). 
For categorical features, we seek to estimate $f(x^*) - f(x^*_\text{base})$, where $x^*_\text{base}$ is $x^*$ with the $j^\text{th}$ entry set to a baseline or counterfactual category, which can be domain-specific or based on the relative frequencies of the categories in the dataset. This represents the amount that $f$ changes, \textit{ceteris paribus}, when the $j^\text{th}$ feature is switched from its baseline category to the current category $x^*_j$.

\paragraph{Uncertainty Intervals} Our second goal is to construct something analogous to a
\textit{confidence interval} $\mathcal{C} \subset \mathbb{R}$,
indicating the level of uncertainty in our point estimate of the feature importance score. A confidence interval at confidence level
$100 \cdot (1-\alpha)\%$ is an interval for which
$100 \cdot (1-\alpha)\%$ of the intervals constructed from repeated
samples contain the true parameter of interest (in our case, the
partial derivatives or function differences of $f$ with respect to
each feature). The most common value of $\alpha$ is $0.05$, resulting
in a confidence level of 95\%.

We will refer to the estimated uncertainty around our explanations 
as ``uncertainty intervals'' rather than ``confidence intervals''.  
We make this distinction to re-enforce that our
uncertainty measures do not correspond to the frequentist definition
of a confidence interval.  This is once again due to the violated
assumption of correct model specification.  It is
known that model misspecification can either increase or decrease the power of a statistical test \cite{powsize}, and that
correcting for the misspecification is often a problem-specific task \cite{ps1,ps2}.  
Thus, we cannot compare the  
%our departure from standard statistical assumptions prevents us from comparing the 
coverage probability of our bootstrap intervals with a pre-specified significance level (as is typically done in frequentist statistics). Instead, we will characterize the efficacy of our method by examining the \emph{trade-off}---in the form of a ``Pareto frontier''---between the coverage probability and interval width.
%Motivated by this, rather than compare the coverage probability of our bootstrap intervals with a pre-specified significance level (as is typically done in frequentist statistics), we will characterize the efficacy of our method by examining the \emph{trade-off}---in the form of a ``Pareto frontier''---between coverage probability and interval width.

%% file: model.tex
\section{Algorithms}
We now describe our algorithms for generating local explanations with
uncertainty quantification, given only a static dataset of input-output
pairs from the model. The full algorithm specifications can be found in the supplementary material.  For simplicity, we focus on explanations based on unregularized polynomial regression. 
%description of the algorithm to unregularized polynomial explainable models.  
However, we describe how our approach can be used with any plug-and-play explainability method, including regularized local models. 
\paragraph{Estimating Feature Importance} 
Our approach to estimating the local feature importance scores of $f$ around $x^*$ is outlined in %Algorithm~\ref{alg:explanation} in 
the supplementary material and can be summarized as follows: 
\begin{enumerate}[itemsep=0pt]
    \item Within the dataset $\mathcal{D}$, identify the $m$ points closest to $x^*$ according to the chosen proximity function $\pi_{x^*}(\cdot)$. Denote this neighborhood around $x^*$ by $\mathcal{N}_{x^*}$. 
    \item Fit a degree-$k$ polynomial, $g$, to the local dataset $\{(z, f(z)) \, : \, z \in \mathcal{N}_{x^*}\} \subseteq \mathcal{D}$. 
    \item Return the partial derivative or local function differences of $g$ with respect to each feature, evaluated at the query point $x^*$.
\end{enumerate}
%\sa{Write out the normal equations for weighted regression. Also maybe cite LOWESS papers.}

The primary hyperparameters of our algorithm are $m$---which controls
the size of the neighborhood around $x^*$---and $k$---which is the
degree of the local polynomial fit to the neighborhood. There is a
trade-off associated with each hyperparameter. Increasing $m$ provides
more data for the regression but may diminish the quality of
the estimate by including points that are further away from $x^*$ and
hence less relevant to its local explanation. 
A ``middle ground'' can be achieved by performing weighted regression in Step 2 above, with weights given by
$\{\pi'_{x^*}(z) \, : \, z \in \mathcal{N}_{x^*}\}$, where
$\pi'_{x^*}(\cdot)$ is some proximity function (not necessarily the
 same as $\pi_{x^*}(\cdot)$).
 % that assigns higher weights to points closer to $x^*$.
In general, $m$ should
scale with the ``density'' of the dataset $\mathcal{D}$ in the
vicinity of $x^*$, since a higher density means there are more points
similar to $x^*$ that can potentially aid the explanation. If the goal is to estimate function differences, the neighborhood size should also take into account the $\delta$ values for each continuous feature. 
Similarly, a larger $k$ allows for a more expressive model (and potentially a better local approximation to $f$) but at the cost of requiring a
larger $m$.\footnote{A polynomial of degree $k$ is uniquely determined by $k+1$ points.} In most instances, we found
$k \leq 4$ to yield good performance. 
We discuss further strategies for
selecting the hyperparameters in Section \ref{sec:setparams}. 

One important feature of our method is that although we use polynomial
regression in Step 2 to construct the explanations, our approach can be used
in a ``plug-and-play'' fashion with any local interpretable model
(e.g., decision trees), not just polynomials. However, we focus on
polynomial models because they allow us to derive theoretical
confidence intervals to use as a baseline in our simulation study (see
Section~\ref{subsec:theoretical}).  Furthermore, the interpretable
model can include regularization terms to reduce model complexity at
the cost of small sample bias.  

Our method of generating local explanations is similar to LIME, with two notable exceptions. First, our method uses only the points provided in the static dataset to construct the explanations, while LIME assumes one has an unrestricted ability to query the underlying model. Second, our explanations are derived from a local polynomial model, whereas LIME assumes the underlying model is locally \textit{linear} around the target point. This is an important distinction: Since LIME assumes model access and can sample points arbitrarily close to the target point, its assumption of local linearity likely holds in most cases. In our fixed-dataset environment, however, we need to endow our local model with more flexibility.

\paragraph{Bootstrap Uncertainty Intervals} The \textit{bootstrap}
\cite{efron1992bootstrap} is a well-known resampling method for
estimating the sampling distribution of any statistic. It is
particularly useful when the sampling distribution is unknown, as in our setting. Our high-level approach is to repeatedly
perform the previously described local regression procedure on \textit{sub-samples} of the
neighborhood around $x^*$ to build a bootstrap distribution of feature
importance scores, from which we derive our uncertainty intervals. Our algorithm
computes the \textit{percentile bootstrap}
\cite{efron1994introduction}, though we note that there are many alternative
methods for constructing confidence intervals from the bootstrap
distribution. The algorithm is fully specified in 
%Algorithm~\ref{alg:bootstrap} in 
the supplementary material and can be summarized as follows: 
\begin{enumerate}[itemsep=0pt]
    \setcounter{enumi}{-1}
    \item As before, let $\mathcal{N}_{x^*}$ be the neighborhood around $x^*$ with $\vert\mathcal{N}_{x^*}\vert = m$, and let  $m' < m$ be an integer denoting the size of each bootstrap sample.
    \item For some $c \in (0,1)$, draw a sample of size $m'=\lfloor{cm}\rfloor$  uniformly at random from $\mathcal{N}_{x^*}$, creating the sub-neighborhood $\mathcal{N}_{x^*}' \subset \mathcal{N}_{x^*}$.
    \item Fit a degree-$k$ polynomial to $\{(z, f(z)) \, : \, z \in \mathcal{N}_{x^*}'\} \subset \mathcal{D}$, and record the estimated feature importance scores. 
    \item Repeat Steps 1 and 2 many times (e.g., $1{,}000$), recording the feature importance scores obtained at every iteration.
    \item For each $j \in \{1,\ldots,d\}$, return the uncertainty interval $[L_j, \, U_j]$ where $L_j$ and $U_j$ are, respectively, the $(100 \cdot \frac{\alpha}{2})^\text{th}$ and $(100 \cdot (1-\frac{\alpha}{2}))^\text{th}$ percentiles of the bootstrapped feature importance scores for feature $j$. 
\end{enumerate}

%% file: results.tex
\section{Simulation Study}\label{sec:experiments}
% \sa{Mention that we focus on tabular data, although in principle our methods could apply to image \& text data too (future work). Also specify which proximity functions we use in the experiments (for neighbor selection \& regression weights). }

% \sa{One can utilize a goodness of fit measure such as $R^2$ to select the hyperparameters ($m,k$) in a more automated fashion.} 

\begin{comment} 
\begin{figure}[t]
  \centering
    \includegraphics[width=0.25\textwidth]{f2}
\caption{Ground truth model for our simulation study, defined in \eqref{eqn:ground_truth}, with $a=b=3$.}
\label{fig:simmod}
\end{figure}
\end{comment} 

\begin{figure}[t]%{.3\textwidth}
  \centering
    \begin{subfigure}{.25\textwidth}
    \centering%\captionsetup[subfigure]{justification=centering}
    \includegraphics[width=\linewidth]{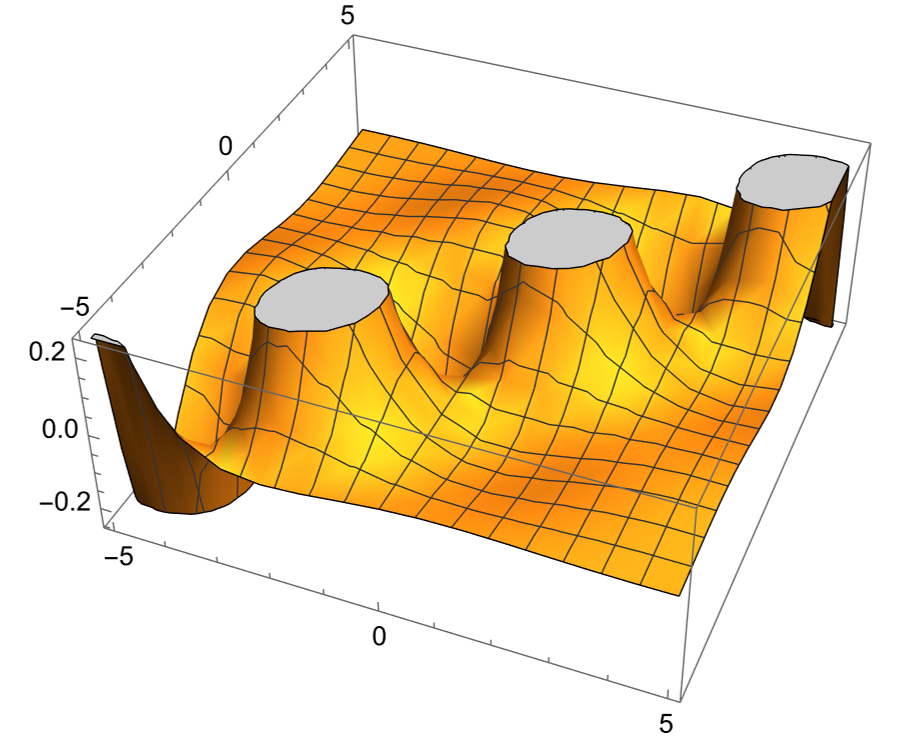}
    \caption{$a=b=1$}
    \label{fig:1}\par\vfill
  \end{subfigure}\hspace{5em}
  \begin{subfigure}{.25\textwidth}
    \centering 
    \includegraphics[width=\linewidth]{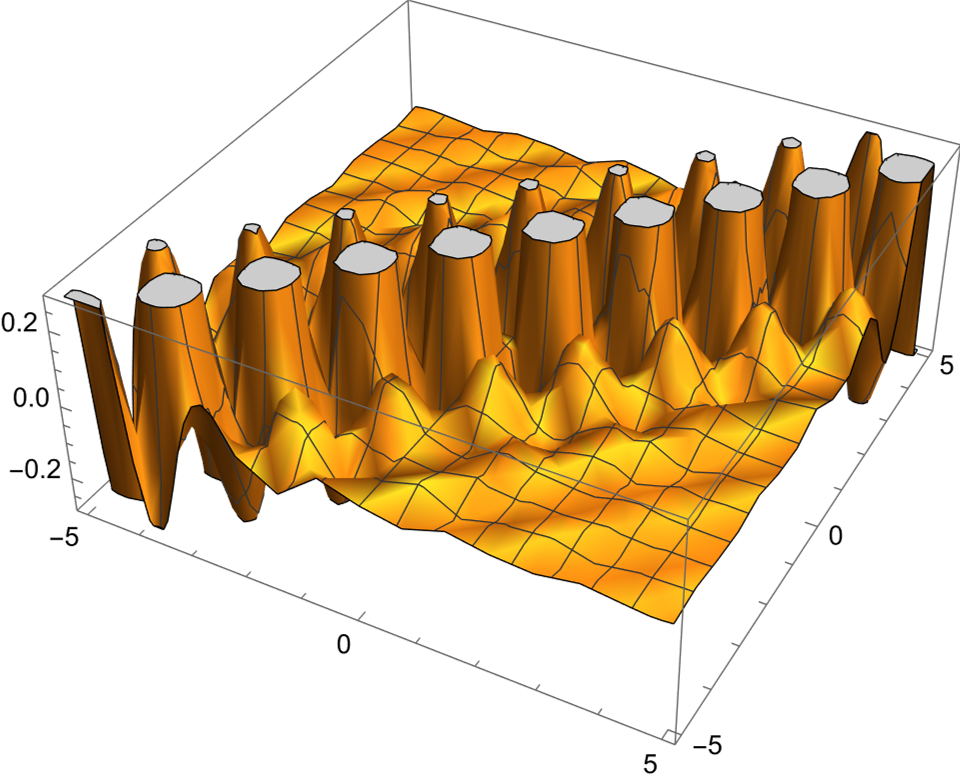}
    \caption{$a=b=3$}
    \label{fig:2}
\end{subfigure}
\caption{Ground truth model for our simulation study, defined in \eqref{eqn:ground_truth}.}
\label{fig:simmod}
\end{figure}

In this section, we demonstrate through Monte Carlo simulations that our non-parametric bootstrap method outperforms the naive theoretical confidence
intervals from classical regression analysis.  Specifically, we show that our approach attains a more favorable Pareto frontier representing the trade-off between interval width and coverage probability.

\subsection{Naive Confidence Intervals}\label{subsec:theoretical}

 To obtain confidence intervals for our explanations, one could naively assume that the sampling distribution of the estimated feature importance score is asymptotically a Gaussian distribution centered around the true importance score. This is true if one is estimating partial derivatives using local polynomial regression and if the following assumptions from classical regression analysis hold: (i) The model is well-specified (it locally matches the true underlying model). (ii) Any error in the fit is due to mean-zero, i.i.d. Gaussian noise. 
 %For partial derivative estimation, this holds true in the classical regression setting where it is assumed that the local model is well-specified (it accurately captures the underlying data-generating process), and that any error in the fit is due to mean-zero, i.i.d. Gaussian noise. 
%In this case, we have the following closed-form expression for the confidence interval of the importance score at confidence level $100 \times (1-\alpha)\%$: 
In this case, the confidence interval of the importance score at confidence level $100 \times (1-\alpha)\%$ is given by the closed-form expression %$\Big[\hat{\theta} - z^* \cdot \text{se(\hat{\theta}), \, \hat{\theta} + z^* \cdot \text{se}(\hat{\theta})\Big].$  
\begin{equation}\label{eqn:theoretical_CI}
    \Big[\hat{\theta} - z^* \cdot \text{se}(\hat{\theta}), \quad \hat{\theta} + z^* \cdot \text{se}(\hat{\theta})\Big].
\end{equation}
Here, $\hat{\theta}$ is the point estimator of the partial derivative, $z^*$ is the z-score corresponding to confidence level $100 \times (1-\alpha)\%$, and $\text{se}(\hat{\theta})$ is the standard error of the sampling distribution. For example, if a $95\%$ confidence level is desired, the resulting bounds are $\hat{\theta} \pm 1.96 \times \text{se}(\hat{\theta})$. %Henceforth, we will refer to \eqref{eqn:theoretical_CI} as the \textit{theoretical confidence interval}, and we will treat this as the baseline approach. 

Since the local polynomial model is linear in the regression coefficients, it follows that the partial derivative of the model with respect to any feature is also linear in the coefficients. Hence, $\text{se}(\hat{\theta}) \triangleq \sqrt{\text{Var}(\hat{\theta})}$ can be determined by computing the variance of a linear combination of the regression coefficients as follows. For simplicity, we consider the one-dimensional case and note that the derivation extends straightforwardly to the case of multiple covariates. Let $g$ be the local polynomial model used in our explanation, which can be expressed as $g(x) = \sum_{\ell=0}^k \hat{\beta}_\ell \cdot x^\ell$. Then the estimated feature importance score is given by $\hat{\theta} = g'(x) = \hat{\beta}^\top v$, where $\hat{\beta} = \begin{bmatrix}\hat{\beta}_1 & \hat{\beta}_2 & \cdots & \hat{\beta}_k\end{bmatrix}^\top$ and $v \in \mathbb{R}^k$ is a vector with $v_\ell = \ell \cdot x^{\ell-1}, \, \ell \in \{1,2,\ldots,k\}$.
It follows that %\text{Var}(\hat{\theta}) = \text{Var}(\hat{\beta}^\top v) = v^\top \Sigma v$,
\begin{equation} 
\text{Var}(\hat{\theta}) = \text{Var}(\hat{\beta}^\top v) = v^\top \Sigma v,
\end{equation}
where $\Sigma \in \mathbb{R}^{k \times k}$ is the variance-covariance matrix of regression coefficients, such that $\Sigma_{ij} = \text{Cov}(\hat{\beta}_i, \hat{\beta}_j)$ for $i,j \in \{1,\ldots, k\}$. In the case of the least squares estimator, we have %$\hat{\beta} = (X^\top X)^{-1}X^\top y$ and $\Sigma = (X^\top X)^{-1}\sigma^2$,
\begin{equation}
    \hat{\beta} = (X^\top X)^{-1}X^\top y, \quad \Sigma = (X^\top X)^{-1}\sigma^2,
\end{equation}
where $\sigma^2$ (the noise variance) can be estimated as %$\hat{\sigma}^2 = \frac{1}{m-d-1}\sum_{z \in \mathcal{N}_{x^*}} \Big(f(z) - g(z)\Big)^2.$
\begin{equation}
    \hat{\sigma}^2 = \frac{1}{m-d-1}\sum_{z \in \mathcal{N}_{x^*}} \Big(f(z) - g(z)\Big)^2.
\end{equation}

%<<<<<<< HEAD
%\subsection{Comparison with Naive Confidence Intervals} 
\subsection{Simulation Results} 

As a ground truth model, we use the function
\begin{align} 
%S(x_1, x_2, a,b) =\sin(a\cdot x_1)\cdot \cos(b\cdot x_2)\cdot \tan(\frac{1}{1+(x_1-x_2)^2}), \label{eqn:ground_truth}
S(x_1, x_2, a,b) =\sin(a \cdot x_1) \cdot \cos(b \cdot x_2) \cdot \tan\Bigg(\frac{1}{1+(x_1-x_2)^2}\Bigg), \label{eqn:ground_truth}
\end{align} 
where $x_1, x_2 \in [-5,5]$ are continuous features and
$a,b\in\{1,2,3\}$ are categorical features.  The model is depicted 
for two different values of $a$ and $b$ in Figure~\ref{fig:simmod}.  This function captures several complex characteristics often found in real-world ML models: 
it is not defined everywhere, it approaches infinity in
some regions, and it is drastically sensitive to changes in the
categorical values.

\begin{figure}
\centering
\begin{subfigure}{.49\textwidth}
  \centering
  \includegraphics[width=\linewidth]{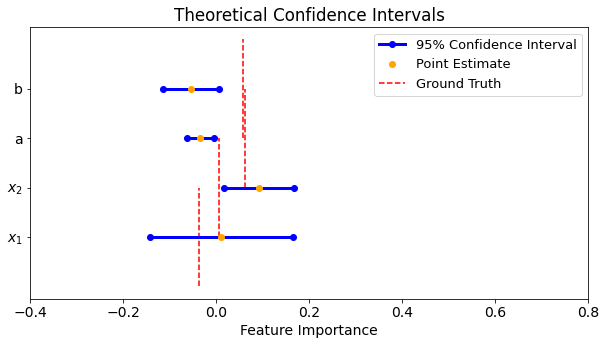}
  \caption{}
  \label{fig:simtheo}
\end{subfigure}
\begin{subfigure}{.49\textwidth}
  \centering
  \includegraphics[width=\linewidth]{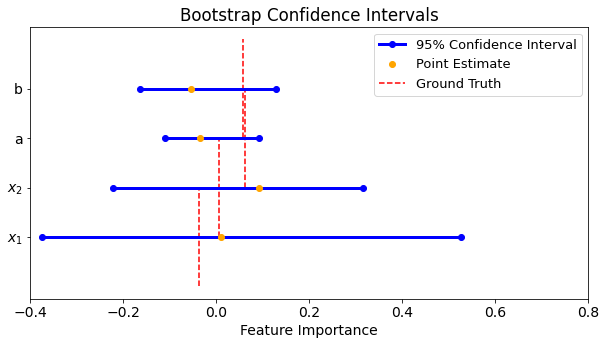}
  \caption{}
  \label{fig:simboot}
\end{subfigure}
\caption{Simulated comparison between the theoretical and bootstrap intervals with $n=2{,}000$. % sampled uniformly from the domain. To construct the explanations, we standardized the continuous features and one-hot encoded the categorical features, then fit a polynomial of degree $k=4$ using weighted least squares regression with the $m=66$ closest points.  For the confidence intervals, we drew $B=500$ bootstrap samples of size $m'=59$, sampled uniformly from the neighborhood of $m=66$ points. 
%Continuous features were standardized and categorical features were one-hot encoded. We fit a polynomial of degree $k=4$ using weighted least squares with the $m=66$ closest points. For the confidence intervals, we drew $B=500$ bootstrap samples of size $m'=59$, sampled uniformly from the neighborhood of $m=66$ points. 
%on an example instance. We created a synthetic dataset of size $n=2{,}000$, with inputs sampled uniformly at random from the domain. To construct the explanations, we standardized the continuous features and one-hot encoded the categorical features, then fit a polynomial of degree $k=4$ using weighted least squares with the $m=66$ closest points.  For the confidence intervals, we drew $B=500$ bootstrap samples of size $m'=59$, sampled uniformly from the neighborhood of $m=66$ points. 
%The yellow dots are point estimates of the feature importance scores and the blue lines are the confidence/uncertainty intervals. 
}
\label{fig:oneexp}
\end{figure}

As a proof of concept, Figure~\ref{fig:oneexp} compares our method to
the naive confidence intervals (using a significance level of 5\% in
both cases) at a randomly selected instance in the simulated data.\footnote{All parameters, including those to generate the figures, can be found in the supplementary material.} The plot shows that our
bootstrap interval captures the true feature importance scores for all
of the features, whereas the naive confidence interval is too narrow
for all but one of the features $(x_1)$.  Of course, in this
particular example, one could improve the performance of the
naive intervals simply by ``stretching'' them (e.g., using a 99\%
confidence level to increase their width).  Thus, it does not make
sense to compare interval widths without controlling for coverage
rate, and vice versa.

\begin{figure}%{.5\textwidth}
\centering
\begin{subfigure}{0.49\textwidth}
    \includegraphics[width=\linewidth]{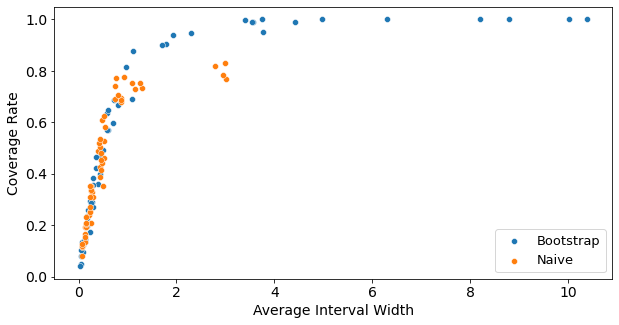}
    \caption{Naive confidence intervals}
    \label{fig:dominance}
\end{subfigure}
\begin{subfigure}{.49\textwidth}
    \includegraphics[width=\linewidth]{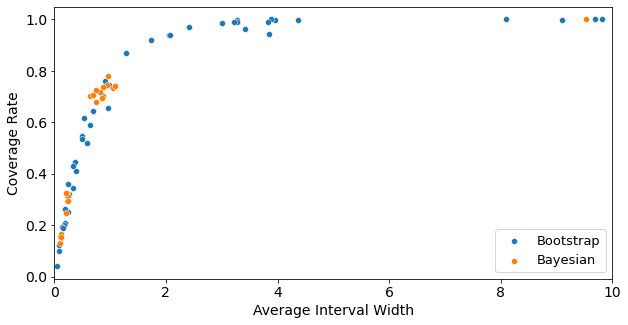}
    \caption{BayesLIME}
    \label{fig:bl}
\end{subfigure}
\caption{Comparison of the Pareto frontiers of the bootstrap uncertainty intervals and other methods.} 
\end{figure}

To ensure a proper comparison between the naive and bootstrap
 intervals, we analyze each method's ``Pareto frontier'' for
variable $x_1$ of the ground truth model.  Specifically, we vary $m$, $m'$ and
$k$, and for each parameter set, we sample $p$ points from the function
\eqref{eqn:ground_truth}.  Using those $p$ samples, we compute the average interval width
and the empirical coverage probability for both the naive
confidence interval and our bootstrap uncertainty interval.  This
yields two sets of tuples---one set corresponding to the naive
intervals, the other to the bootstrap----of the form
$(\texttt{Average Interval Width}, \, \texttt{Coverage Rate})$.  To
visualize the Pareto frontier, we plot these points in the two-dimensional plane as shown in Figure \ref{fig:dominance}.

% Figure~\ref{fig:dominance} shows that when we
% calibrate our hyperparameters such that the bootstrap method has at
% least as high a coverage probability as the theoretical confidence
% intervals, the bootstrap confidence intervals are narrower, thus
% demonstrating more certainty in the estimate. Further experimental
% details can be found in the supplementary material.

Figure \ref{fig:dominance} shows that our bootstrap method weakly
dominates the naive theoretical intervals.  For low
coverage rates, the naive intervals and the bootstrap intervals exhibit
the same trade-off between coverage rate and average interval width.
However, around a coverage rate of $0.8$, the naive
intervals begin expanding with little gain in coverage rate, whereas
the bootstrap method continues to increase its coverage rate as the
interval widens.  The intuition behind this result is two-fold.
First, our bootstrap intervals are not centered around the
point estimate, which allows them to have a higher coverage rate for a
given interval width.  Second, both methods have higher coverage
rates when the local neighborhood is of moderate size % (see section
% \ref{sec:setparams} for more detail on how coverage rates depend on
% the hyper-parameters)
but the polynomial degree is relatively high.  This
means that the naive approach has very few degrees of
freedom, resulting in intervals that are not wide enough to
cover the true parameter unless the point estimate is sufficiently
close to the ground truth.

\subsection{Choosing Hyperparameters}
\label{sec:setparams}

\begin{figure}[h]%{.5\textwidth}
\centering
\includegraphics[width=0.6\textwidth]{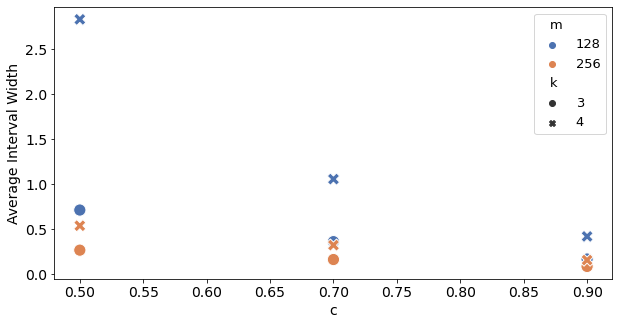}
\caption{Average width of bootstrap uncertainty intervals as a function of $m, k, c$.}
\label{fig:setparams}
\end{figure}

Figure \ref{fig:dominance} also  provides important guidance on how to
choose hyperparameters.  Since, in general, the points lie on the
Pareto frontier (aberrations are likely due to noise in the Monte
Carlo procedure), a practitioner can specify a desired average interval width
and set the hyperparameters to meet such a specification.  While 
it would be impossible to know the coverage
rate without a ground truth, the fact that all points lie on the Pareto frontier guarantee
that no choice of hyperparameters is inefficient.  That is, for a
given choice of hyperparameters yielding a particular average interval width,
there is not another set of hyperparameters yielding the same
interval width with a higher coverage rate.

Figure \ref{fig:setparams} shows how to adjust the hyperparameters to
achieve a desired interval width.
Generally speaking, increasing
$m'$ decreases the interval width, as illustrated by the downward
tendency of the points.  Increasing $m$ also decreases the interval
width, as illustrated by the higher positioning of the blue points compared
to the orange points.  Finally, increasing $k$ increases the interval width, as
illustrated by the higher positioning of the $\texttt{x}$ marks compared to the
$\texttt{o}$ marks.  These patterns make intuitive
sense: the larger the fraction of points $c$ used to compute the
 intervals, the larger the ``overlap'' between the bootstrap samples, and thus the smaller the resulting interval width.  
 The larger the local neighborhood ($m$), the less likely it is that the polynomial model captures the true ML model over the entire region; and if the true model varies significantly more than the flexibility of the polynomial allows, the polynomial coefficients will attenuate toward $0$ and the resulting bootstrap interval width will decrease.\footnote{To further drive home this intuition, suppose the true underlying model is a sine wave and the local polynomial is a line. The sine can be well-approximated by the line over small intervals (e.g., less than a quarter-period). However, as the interval expands, the approximation worsens and the fitted polynomial converges to a horizontal line (assuming un-weighted regression is performed). This intuition roughly extends to our ground-truth model \eqref{eqn:ground_truth}, which exhibits similar periodic behavior, as illustrated in Figure~\ref{fig:simmod}. } 
 %the smaller the likelihood that the polynomial model fits the true underlying ML model, which causes the parameters to attenuate toward $0$ for each bootstrap sample and the resulting interval width to decrease.  
Finally, higher-degree polynomials exhibit higher variance and have fewer
degrees of freedom; thus, the bootstrap distribution is wider for
larger $k$.  

\subsection{Comparison with BayesLIME}

BayesLIME \cite{reliable1} is an existing method for quantifying the uncertainty of 
model-agnostic explanations.  In this section, we compare our bootstrap-based
explanations and uncertainty measures with those of BayesLIME.  We
emphasize, however, that the environment for which BayesLIME was
developed is different from ours in that BayesLIME assumes access to the
underlying machine learning model 
and the ability to take an arbitrarily large number of samples from it.  
This inhibits a perfect
apples-to-apples comparison between our bootstrap method and
BayesLIME.  
However, we take the following measures to increase the fairness of the comparison:
(i) We adapt BayesLIME to only %be able to generate
leverage samples from the same fixed dataset available to our bootstrap method.
(ii) To ensure that BayesLIME is not handicapped by its assumption of linearity,
we endow BayesLIME with the ability to fit higher-order polynomials.
The purpose of this exercise is not to show that our bootstrap
method is unambiguously better than BayesLIME, but instead to show that when
the assumption of model access is loosened, the leading methods that
rely on model access may no longer be appropriate.

\begin{comment}
\begin{figure}%{.5\textwidth}
\centering
\includegraphics[width=0.6\textwidth]{bayescompare}
\caption{Comparison of the Pareto frontiers of the BayesLIME and bootstrap uncertainty intervals.}
\label{fig:bl}
\end{figure}
\end{comment}

  Figure
\ref{fig:bl} once again plots the Pareto frontier representing the trade-off between interval
width and coverage rate.  As in the case of the naive uncertainty
intervals (Figure~\ref{fig:dominance}), BayesLIME does not acheive a coverage rate above $0.8$---unless the intervals are made impractically large.  While there may be intermediate parameter sets in which BayesLIME achieves a coverage
rate greater than $0.8$, its greater sensitivity to hyperparameter
selection inhibits its practicality.  On the other hand, our bootstrap
method traces a relatively smooth curve through the space and enables
practitioners to better fine-tune their desired interval width.  This
provides evidence that without model access, our bootstrap method can
achieve higher coverage rates than BayesLIME with more flexibility in
setting the hyperparameters.

%% file: realdata.tex
\section{Experiments with Real-World Data and Models}\label{sec:realdata}

\begin{figure}
\centering
\begin{subfigure}{.5\textwidth}
  \centering
  \includegraphics[width=\linewidth]{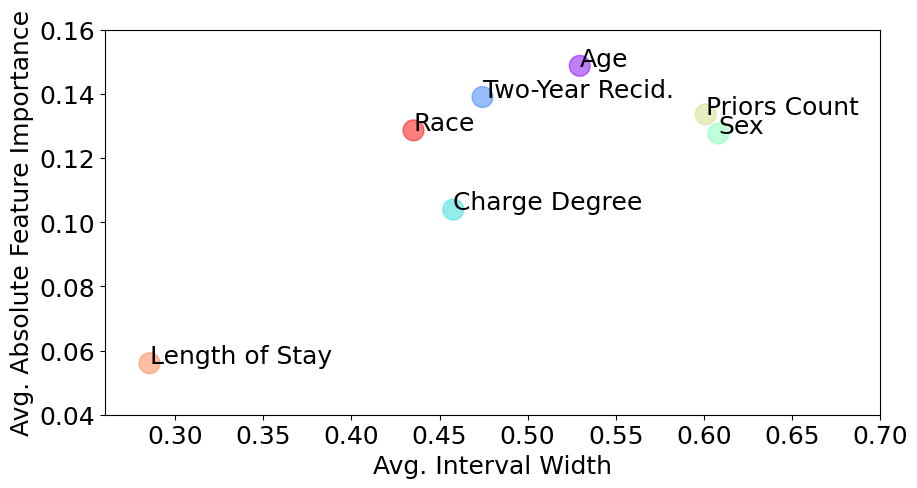}
  \caption{COMPAS + random forest}
  \label{fig:compas}
\end{subfigure}%=
\begin{subfigure}{.5\textwidth}
  \centering
  \includegraphics[width=\linewidth]{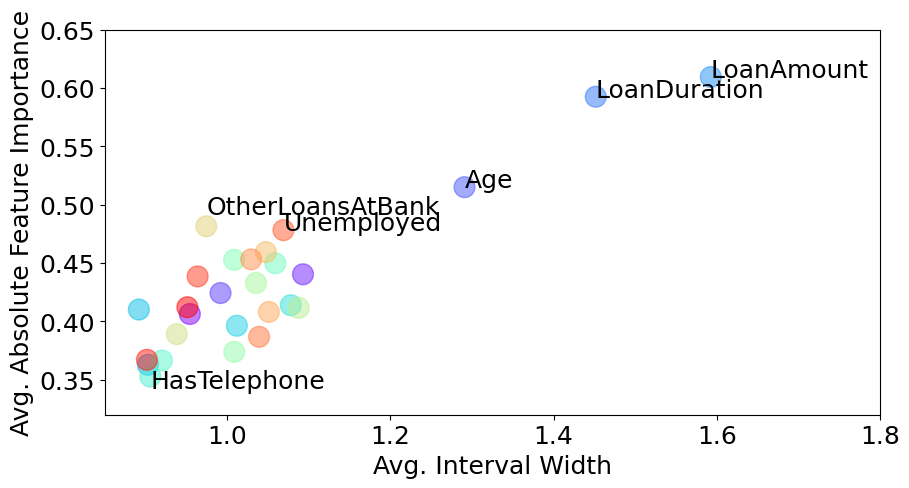}
  \caption{German Credit + random forest}
  \label{fig:german}
\end{subfigure}
\begin{subfigure}{.5\textwidth}
  \centering
  %\vspace{1em}
  \includegraphics[width=\linewidth]{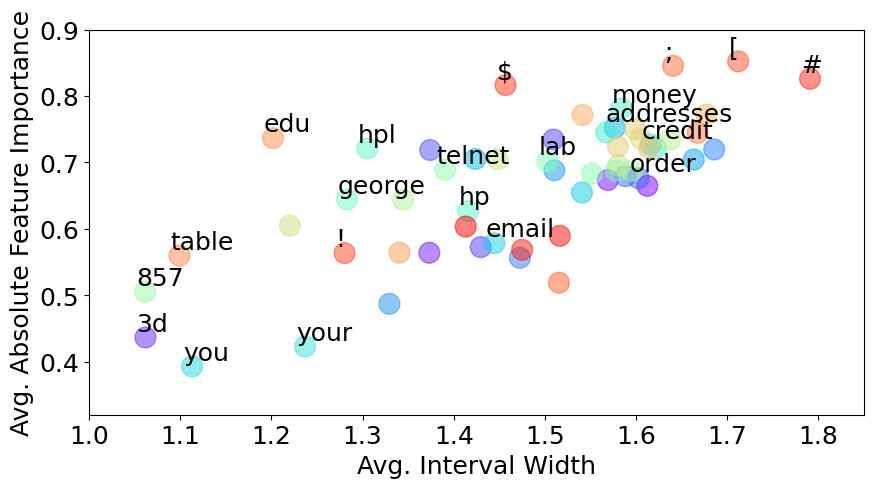}
  \caption{Spambase + SAINT neural network}
  \label{fig:spambase}
\end{subfigure}
\caption{Visual summary of feature importance scores and corresponding uncertainty measures for models trained on three different datasets.}
\label{fig:realdata}
\end{figure}

We further demonstrate the capabilities of our method by applying it to models trained on three well-known tabular datasets: the German Credit and Spambase datasets from the UCI machine learning repository \cite{dua2017uci}, and the COMPAS recidivism dataset \cite{angwin2016machine}. 

\subsection{Random Forest Classifiers}\label{subsec:rf_experiments}
For the German Credit and COMPAS datasets, we trained a random forest classifier with $100$ estimators on a random 80-20 train-test split. We treated the resulting classifier as a black-box model and collected its predictions on the entire dataset to form our static dataset of model inputs and outputs. Due to the step-function-like behavior of the random forest models, we estimated the function differences with our bootstrap method instead of the partial derivatives. Additionally, before performing local regression, we standardized the continuous features, one-hot encoded the categorical features, and transformed the response (i.e., the probabilities predicted by the model) to log-odds. (However, we converted log-odds back to probabilities when estimating the feature importance scores and uncertainty intervals.) The parameters we used were $k=2, \, m=40, \, c=0.9$ and $k=2, \, m=150, \, c=0.667$ for the German Credit and COMPAS datasets, respectively. 

We generated an explanation of model behavior for every instance in the test set and computed the following quantities for each feature: (i) the average absolute feature importance score (i.e., the magnitude of the estimated importance assigned to each feature, averaged over all test instances); and (ii) the average width of the uncertainty intervals.  Figure~\ref{fig:realdata} plots the resulting $(\texttt{width},\, \texttt{importance})$ tuples for each feature. These plots provide a high-level view of the average influence of each feature on the model's predictions, along with the corresponding uncertainty that arises from using a finite sample to estimate these quantities. For example, Figure~\ref{fig:compas} shows that the $\texttt{Priors Count}$ and $\texttt{Sex}$ features have a high average importance but are also accompanied by high uncertainty. On the other hand, $\texttt{Length of Stay}$ has a lower average importance but also lower uncertainty. Finally, the $\texttt{Race}$ feature has a relatively high average importance along with a moderate level of uncertainty. 
Similar observations can be made about Figure~\ref{fig:german}. We see that the $\texttt{OtherLoansAtBank}$ and $\texttt{Unemployed}$ features have both high average importance and low uncertainty, whereas $\texttt{LoanDuration}$ and $\texttt{LoanAmount}$ have higher uncertainty, and  $\texttt{HasTelephone}$ has both low average importance and low uncertainty. In practice, after making such observations, one might further investigate the model for issues related to bias and fairness, or attempt to improve the fidelity of the explanation by collecting more data based on the highest-uncertainty features.

\subsection{Deep Neural Network}
To demonstrate the utility of our method for more complex models, we used SAINT \cite{somepalli2021saint}, which is an attention-based neural network for tabular data recently shown to outperform more traditional tabular learning algorithms on a variety of supervised and semi-supervised tasks. We trained the model with its default parameters\footnote{\url{https://github.com/somepago/saint}} on the Spambase dataset. Each row in the dataset is an email which is labeled as either spam or non-spam. Each email is represented by 57 features, most of which measure the frequencies of different words or characters in the message. As with the random forest classifiers, we treated the SAINT model as a black box and collected its inputs and outputs on the dataset. We also estimated the function differences of the model with respect to each feature, and we applied the same standardization and encoding procedures specified in Section~\ref{subsec:rf_experiments} before performing polynomial regression. The parameters we used were $k=2, \, m=150, \, c=0.9$.

Figure~\ref{fig:spambase} plots, for each feature, the absolute feature importance score and bootstrap interval width, averaged over all instances in the validation set. Somewhat unsurprisingly, keywords and characters such as $\texttt{money}, \texttt{addresses}, \texttt{credit}, \texttt{order}, \texttt{\$}$, and $\texttt{\#}$ had high average absolute importance scores. However, they also had relatively large interval widths. One possible explanation is that these words and characters commonly appear in both spam and non-spam emails (e.g., confirmations for online shopping orders). Therefore, it is unclear whether these features are truly as indicative of the spam label as their importance scores would suggest. Examples of high-importance, mid-interval-width features include $\texttt{edu}, \texttt{hpl},$ and  $\texttt{george}$, the presence of which likely increase the probability that a message is non-spam.\footnote{The Spambase dataset was initially compiled by Hewlett-Packard Labs, which is why the acronym $\texttt{hpl}$ is a feature in the dataset. One of the dataset's creators was George Forman, hence the $\texttt{george}$ feature.} Lastly, features such as $\texttt{you}$ and $\texttt{your}$ had relatively small importance scores and interval widths, matching our intuition that such commonly used words would not be particularly influential features in a spam classifier.

%% file: future.tex
\section{Conclusions and Open Problems}
We proposed a method for generating local explanations from a fixed dataset of model queries, obviating the need for direct model access. An essential component of our method is a non-parametric bootstrapping technique for
quantifying the uncertainty in the explanations, motivated by the fact that our setting does not fit neatly into the standard statistical estimation paradigm. Through simulation studies, we showed that our
bootstrap confidence intervals outperform (i) a theoretical approach
from frequentist statistics that makes strong distributional
assumptions about the estimated feature importance scores; and (ii) a
Bayesian method for quantifying the uncertainty in explanations
\cite{reliable1}.
% assumes the sampling distribution of the estimated feature
% importance score is a Gaussian centered around the true score.
Lastly, we applied our method to three real-world datasets and demonstrated its ability to simultaneously yield insights into the behavior of the black-box model as well as the inherent uncertainty of our explanations due to the fixed nature of the query data. 

An important use-case we omitted is that of image and language
models. In theory, our method can be applied as-is to both types of models: we can fit a local interpretable model, compute derivatives or finite differences, and generate bootstrap uncertainty intervals.  However, without the ability to query the model directly, computing a local model is likely infeasible.  For example, when applying LIME to image data, the importance of a super-pixel is estimated by ``blacking out'' a region in the image and comparing the model's output on the blacked-out image with its
output on the original image.  To compute the explanations (and thus uncertainty intervals), our method would require a specialized dataset consisting of complete images as well as blacked-out images. An alternative approach would be to ``steal'' the model by training another model on the given dataset, using the stolen model as a surrogate for the true model, then fitting an interpretable model to the surrogate.  However, training the (possibly large and complex) surrogate model may be too computationally intensive to be practical. Furthermore, if the dataset is not sufficiently dense around the input point of interest, the output of the surrogate model will not be aligned with the output of the true model.  For these reasons, building model explanations and uncertainty intervals for language and image models from low-density datasets remains an open problem.

%Furthermore, our original motivation for using a non-parametric
%bootstrap to construct uncertainty intervals was that task of creating
%explanations does not fit into the standard statistical estimation
%paradigm.  Therefore, an alternative route forward is to instead
%borrow tools from approximation theory \cite{approx1} that abandons
%the statistical paradigm altogether.  

%% file: supp.tex
\appendix

\section{Algorithms and Time Complexity}

\begin{algorithm}
\caption{Estimating feature importance scores via local polynomial regression}
\label{alg:explanation}
\begin{algorithmic}[1]
    \STATE \textbf{Input: } Instance $x^* \in \mathbb{R}^d$, dataset $\mathcal{D}$, polynomial degree $k$, neighborhood size $m$, proximity function $\pi_{x^*}(\cdot)$ 
    \STATE \textbf{Initialize: } $Z \in \mathbb{R}^d$
    \STATE $\mathcal{N}_{x^*} \leftarrow m$ closest points to $x^*$ in $\mathcal{D}$ according to $\pi_{x^*}(\cdot)$ 
    %\STATE $y \leftarrow \begin{bmatrix}f(z), \, z \in \mathcal{N}_{x^*}\end{bmatrix}$%: labels of points in $\mathcal{N}_{x^*}$
    \STATE $y \leftarrow$ labels of points in $\mathcal{N}_{x^*}$, taken from $\mathcal{D}$
    \STATE $X \leftarrow $ design matrix of points in $\mathcal{N}_{x^*}$, transformed to polynomial features of degree $k$ with interaction terms
    \STATE $\hat{\beta} \leftarrow (X^\top X)^{-1} X^\top y$
    \STATE $g \leftarrow $ degree-$k$ polynomial with interaction terms, parameterized by $\hat{\beta}$
    \FOR{$j=1$ to $d$}
        \IF{feature $j$ is continuous}
            \STATE $Z_{j} \leftarrow \frac{\partial g}{\partial x_j}(x^*)$ or $g(x^*_+) - g(x^*_-)$
        \ELSIF{feature $j$ is categorical} 
            \STATE $Z_{j} \leftarrow g(x^*) - g(x^*_\text{base})$
        \ENDIF 
    \ENDFOR 
    \STATE \textbf{return} $Z$
\end{algorithmic}
\end{algorithm} 

\begin{algorithm}
\caption{Constructing bootstrap uncertainty intervals}
\label{alg:bootstrap}
\begin{algorithmic}[1]
    \STATE \textbf{Input: } Instance $x^* \in \mathbb{R}^d$, dataset $\mathcal{D}$, polynomial degree $k$, neighborhood size $m$, number of bootstrap samples $B$, bootstrap neighborhood proportion $c$, significance level $\alpha \in [0,1]$ 
    \STATE \textbf{Initialize: } $Z \in \mathbb{R}^{B \times d}$
    \STATE $\mathcal{N}_{x^*} \leftarrow m$ closest points to $x^*$ in $\mathcal{D}$ according to $\pi_{x^*}(\cdot)$ 
    \FOR{$i=1$ to $B$}
        \STATE $\mathcal{N}_{x^*}' \leftarrow \lfloor cm \rfloor$ points sampled uniformly from $\mathcal{N}_{x^*}$
        \STATE $y \leftarrow$ labels of points in $\mathcal{N}_{x^*}'$, taken from $\mathcal{D}$
        \STATE $X \leftarrow $ design matrix of points in $\mathcal{N}_{x^*}'$, transformed to polynomial features of degree $k$ with interaction terms
        \STATE $\hat{\beta} \leftarrow (X^\top X)^{-1}X^\top y$ 
        \STATE $g \leftarrow $ degree-$k$ polynomial with interaction terms, parameterized by $\hat{\beta}$ 
        \FOR{$j=1$ to $d$}
            \IF{feature $j$ is continuous}
                \STATE $Z_{ij} \leftarrow \frac{\partial g}{\partial x_j}(x^*)$ or $g(x^*_+) - g(x^*_-)$
            \ELSIF{feature $j$ is categorical} 
                \STATE $Z_{ij} \leftarrow g(x^*) - g(x^*_\text{base})$
            \ENDIF 
        \ENDFOR 
    \ENDFOR 
    \FOR{$j=1$ to $d$}
        \STATE $L_j \leftarrow (100 \cdot \frac{\alpha}{2})^\text{th}$ percentile of $Z_{:,j}$
        \STATE $U_j \leftarrow (100 \cdot (1-\frac{\alpha}{2}))^\text{th}$ percentile of $Z_{:,j}$
    \ENDFOR 
    \STATE \textbf{return } $[L_j, \, U_j]$ for each $j \in [d]$
\end{algorithmic}
\end{algorithm}

\paragraph{Time Complexity}
The complexity of constructing the explanation itself (Algorithm~\ref{alg:explanation}) is $O(dn + n\log n + mq^2 + q^3)$ using un-weighted regression, where $q$ is the number of covariates used to fit the polynomial model $g$ (which depends on its degree $k$, whether interaction terms are included, how the categorical features are encoded, etc.).  The $dn + n\log n$ term is due to the computations for $\mathcal{N}_{x^*}$, assuming that sorting takes $O(n\log n)$ time and that $\pi_{x^*}(\cdot)$ can be computed in $O(d)$ time. 
%the sorting involved in computing $\mathcal{N}_{x^*}$, assuming that $\pi_{x^*}(\cdot)$ can be computed in $O(d)$ time. 
Fitting the local polynomial requires $O(mq^2 + q^3)$ time, which can be straightforwardly obtained using the  well-known facts that (i) the complexity of a standard matrix product $AB$, where $A \in \mathbb{R}^{a \times b}, B \in \mathbb{R}^{b \times c}$, is $O(abc)$; and (ii) the complexity of matrix inversion $A^{-1}$, where $A \in \mathbb{R}^{a \times a}$, is $O(a^3)$. If weighted regression is used, the resulting complexity is $O(dn + n\log n + m^2q + mq^2 + q^3)$. The complexities of the non-weighted and weighted versions of our bootstrap algorithm (Algorithm~\ref{alg:bootstrap}) can be determined in a similar manner and are given by $O(dn + n\log n + B(m'q^2 + q^3))$ and $O(dn + n\log n + B(m'^2q + m'q^2 + q^3))$, respectively.

\section{Additional Experimental Details}

\paragraph{Compute Resources} The experiments were run on an Apple M1 machine with 16 GB of RAM and the macOS Monterey operating system. 

\paragraph{Proximity Function}
We used the following procedure to select a neighborhood of $m$ points around the input instance $x^*$. First, we sort the points in the dataset by the Euclidean distance between their continuous features and those of $x^*$. 
Within this ordering, we then select the first $m$ points that provide an equal representation of the baseline class and the query class for each categorical feature. We enforce this balanced representation to ensure that for each categorical feature, the function difference with respect to the baseline class can be estimated accurately. Further implementation details can be found in the supplemental code. 

\paragraph{Weighted Regression}
In the experiments where we perform weighted local regression, the weight assigned to the $i^\text{th}$ sample in the dataset is given by 
\[w_i = \frac{1-(\phi_i - \min(\phi))}{\max(\phi)- \min(\phi)}, \]
where $\phi \in \mathbb{R}^n$ is the vector of proximity values between the input instance $x^*$ and all other points in the dataset: $\phi_i = \pi_{x^*}(x^{(i)})$. 

\paragraph{Experiment and Figure Parameters} 
Below we specify the parameters and other settings used to generate Figures~\ref{fig:oneexp} through \ref{fig:setparams}:
\begin{itemize}
\item Figures \ref{fig:simtheo} and \ref{fig:simboot} were generated
  using a synthetic dataset of size $n=2{,}000$ sampled uniformly from
  the domain. To construct the explanations, we standardized the
  continuous features and one-hot encoded the categorical features,
  then fit a polynomial of degree $k=4$ using weighted least squares
  regression with the $m=66$ closest points.  For the confidence
  intervals, we drew $B=500$ bootstrap samples of size $m'=59$,
  sampled uniformly from the neighborhood of $m=66$ points.
\item Figure \ref{fig:dominance} was generated by sweeping over all combinations
  of $k\in\{1,2,3,4\}$, $m\in \{32,64,128,256\}$, and setting
$m'$ such that $m' = \lfloor{cm}\rfloor$ with $c\in\{.3,.5,.7,.9\}$.
  Throughout, $n=2{,}000$ and $p=250$, but qualitatively similar results
  held for $n \in\{1000,2000,3000,4000,5000\}$.
\item Figure \ref{fig:setparams} was generated by setting $n=2{,}000$ and $p=250$, but
  qualitatively similar results held for $n \in\{1000,2000,3000,4000,5000\}$.
  \item Figure \ref{fig:bl} was generated by sweeping over all combinations
  of $k\in\{2,3,4\}$, $m\in \{64,128,256\}$, and setting
$m'$ such that $m' = \lfloor{cm}\rfloor$ with $c\in\{.3,.5,.7,.9\}$.
  Throughout, $n=2{,}000$ and $p=250$, but qualitatively similar results
  held for $n \in\{1000,2000,4000,5000\}$.
\end{itemize}

%% file: main.bbl
\begin{thebibliography}{}

\bibitem[Adadi and Berrada, 2018]{surv4}
Adadi, A. and Berrada, M. (2018).
\newblock Peeking inside the black-box: a survey on explainable artificial
  intelligence ({XAI}).
\newblock {\em IEEE Access}, 6:52138--52160.

\bibitem[Agarwal et~al., 2022a]{robust3}
Agarwal, C., Johnson, N., Pawelczyk, M., Krishna, S., Saxena, E., Zitnik, M.,
  and Lakkaraju, H. (2022a).
\newblock Rethinking stability for attribution-based explanations.
\newblock {\em arXiv preprint arXiv:2203.06877}.

\bibitem[Agarwal et~al., 2022b]{robust2}
Agarwal, C., Zitnik, M., and Lakkaraju, H. (2022b).
\newblock Probing {GNN} explainers: A rigorous theoretical and empirical
  analysis of {GNN} explanation methods.
\newblock In {\em International Conference on Artificial Intelligence and
  Statistics}, pages 8969--8996. PMLR.

\bibitem[Alvarez-Melis and Jaakkola, 2018]{robust1}
Alvarez-Melis, D. and Jaakkola, T.~S. (2018).
\newblock On the robustness of interpretability methods.
\newblock {\em arXiv preprint arXiv:1806.08049}.

\bibitem[Ancona et~al., 2019]{ancona2019gradient}
Ancona, M., Ceolini, E., {\"O}ztireli, C., and Gross, M. (2019).
\newblock Gradient-based attribution methods.
\newblock In {\em Explainable {AI}: Interpreting, Explaining and Visualizing
  Deep Learning}, pages 169--191. Springer.

\bibitem[Angwin et~al., 2016]{angwin2016machine}
Angwin, J., Larson, J., Mattu, S., and Kirchner, L. (2016).
\newblock Machine bias.
\newblock In {\em Ethics of Data and Analytics}, pages 254--264. Auerbach
  Publications.

\bibitem[Baehrens et~al., 2010]{baehrens2010explain}
Baehrens, D., Schroeter, T., Harmeling, S., Kawanabe, M., Hansen, K., and
  M{\"u}ller, K.-R. (2010).
\newblock How to explain individual classification decisions.
\newblock {\em The Journal of Machine Learning Research}, 11:1803--1831.

\bibitem[Bartlett and Hughes, 2020]{ps2}
Bartlett, J.~W. and Hughes, R.~A. (2020).
\newblock Bootstrap inference for multiple imputation under uncongeniality and
  misspecification.
\newblock {\em Statistical Methods in Medical Research}, 29(12):3533--3546.

\bibitem[Bracke et~al., 2019]{loan1}
Bracke, P., Datta, A., Jung, C., and Sen, S. (2019).
\newblock Machine learning explainability in finance: an application to default
  risk analysis.

\bibitem[Bykov et~al., 2020]{howmuch}
Bykov, K., H{\"o}hne, M. M.-C., M{\"u}ller, K.-R., Nakajima, S., and Kloft, M.
  (2020).
\newblock How much can {I} trust you?--quantifying uncertainties in explaining
  neural networks.
\newblock {\em arXiv preprint arXiv:2006.09000}.

\bibitem[Do{\v{s}}ilovi{\'c} et~al., 2018]{surv5}
Do{\v{s}}ilovi{\'c}, F.~K., Br{\v{c}}i{\'c}, M., and Hlupi{\'c}, N. (2018).
\newblock Explainable artificial intelligence: A survey.
\newblock In {\em 2018 41st International Convention on Information and
  Communication Technology, Electronics and Microelectronics (MIPRO)}, pages
  0210--0215. IEEE.

\bibitem[Dua et~al., 2017]{dua2017uci}
Dua, D., Graff, C., et~al. (2017).
\newblock {UCI} machine learning repository.

\bibitem[Efron, 1992]{efron1992bootstrap}
Efron, B. (1992).
\newblock Bootstrap methods: Another look at the jackknife.
\newblock In {\em Breakthroughs in Statistics}, pages 569--593. Springer.

\bibitem[Efron and Tibshirani, 1994]{efron1994introduction}
Efron, B. and Tibshirani, R.~J. (1994).
\newblock {\em An Introduction to the Bootstrap}.
\newblock CRC press.

\bibitem[Gilpin et~al., 2021]{auton1}
Gilpin, L.~H., Penubarthi, V., and Kagal, L. (2021).
\newblock Explaining multimodal errors in autonomous vehicles.
\newblock In {\em 2021 IEEE 8th International Conference on Data Science and
  Advanced Analytics (DSAA)}, pages 1--10.

\bibitem[Goodman and Flaxman, 2017]{goodman2017european}
Goodman, B. and Flaxman, S. (2017).
\newblock European union regulations on algorithmic decision-making and a
  “right to explanation”.
\newblock {\em AI Magazine}, 38(3):50--57.

\bibitem[Gulum et~al., 2021]{cancer}
Gulum, M.~A., Trombley, C.~M., and Kantardzic, M. (2021).
\newblock A review of explainable deep learning cancer detection models in
  medical imaging.
\newblock {\em Applied Sciences}, 11(10):4573.

\bibitem[Hama et~al., 2022]{art2}
Hama, N., Mase, M., and Owen, A.~B. (2022).
\newblock Model free variable importance for high dimensional data.
\newblock {\em arXiv preprint arXiv:2211.08414}.

\bibitem[Hooker et~al., 2019]{roar}
Hooker, S., Erhan, D., Kindermans, P.-J., and Kim, B. (2019).
\newblock A benchmark for interpretability methods in deep neural networks.
\newblock {\em Advances in Neural Information Processing Systems}, 32.

\bibitem[Iadarola et~al., 2021]{malware}
Iadarola, G., Martinelli, F., Mercaldo, F., and Santone, A. (2021).
\newblock Towards an interpretable deep learning model for mobile malware
  detection and family identification.
\newblock {\em Computers \& Security}, 105:102198.

\bibitem[Kim and Shin, 2021]{loan2}
Kim, D.-s. and Shin, S. (2021).
\newblock The economic explainability of machine learning and standard
  econometric models-an application to the {US} mortgage default risk.
\newblock {\em International Journal of Strategic Property Management},
  25(5):396--412.

\bibitem[Linardatos et~al., 2020]{surv3}
Linardatos, P., Papastefanopoulos, V., and Kotsiantis, S. (2020).
\newblock Explainable {AI}: A review of machine learning interpretability
  methods.
\newblock {\em Entropy}, 23(1):18.

\bibitem[Liti{\`e}re et~al., 2007]{powsize}
Liti{\`e}re, S., Alonso, A., and Molenberghs, G. (2007).
\newblock Type {I} and type {II} error under random-effects misspecification in
  generalized linear mixed models.
\newblock {\em Biometrics}, 63(4):1038--1044.

\bibitem[Lundberg and Lee, 2017]{lundberg2017unified}
Lundberg, S.~M. and Lee, S.-I. (2017).
\newblock A unified approach to interpreting model predictions.
\newblock {\em Advances in Neural Information Processing Systems}, 30.

\bibitem[Mase et~al., 2021]{art1}
Mase, M., Owen, A.~B., and Seiler, B.~B. (2021).
\newblock Cohort shapley value for algorithmic fairness.
\newblock {\em arXiv preprint arXiv:2105.07168}.

\bibitem[Mohseni et~al., 2021]{surv2}
Mohseni, S., Zarei, N., and Ragan, E.~D. (2021).
\newblock A multidisciplinary survey and framework for design and evaluation of
  explainable {AI} systems.
\newblock {\em ACM Transactions on Interactive Intelligent Systems (TiiS)},
  11(3-4):1--45.

\bibitem[Nori et~al., 2019]{ebm}
Nori, H., Jenkins, S., Koch, P., and Caruana, R. (2019).
\newblock Interpret{ML}: A unified framework for machine learning
  interpretability.
\newblock {\em arXiv preprint arXiv:1909.09223}.

\bibitem[Patro et~al., 2019]{ucam}
Patro, B.~N., Lunayach, M., Patel, S., and Namboodiri, V.~P. (2019).
\newblock U-cam: Visual explanation using uncertainty based class activation
  maps.
\newblock In {\em Proceedings of the IEEE/CVF International Conference on
  Computer Vision}, pages 7444--7453.

\bibitem[Ribeiro et~al., 2016]{lime1}
Ribeiro, M.~T., Singh, S., and Guestrin, C. (2016).
\newblock ``{W}hy should {I} trust you?'' {E}xplaining the predictions of any
  classifier.
\newblock In {\em Proceedings of the 22nd ACM SIGKDD International Conference
  on Knowledge Discovery and Data Mining}, pages 1135--1144.

\bibitem[Riedl, 2019]{riedl2019human}
Riedl, M.~O. (2019).
\newblock Human-centered artificial intelligence and machine learning.
\newblock {\em Human Behavior and Emerging Technologies}, 1(1):33--36.

\bibitem[Schulz et~al., 2021]{ordinal}
Schulz, J., Poyiadzi, R., and Santos-Rodriguez, R. (2021).
\newblock Uncertainty quantification of surrogate explanations: an ordinal
  consensus approach.
\newblock {\em arXiv preprint arXiv:2111.09121}.

\bibitem[Schwab and Karlen, 2019]{cxplain}
Schwab, P. and Karlen, W. (2019).
\newblock {CXPlain}: Causal explanations for model interpretation under
  uncertainty.
\newblock {\em Advances in Neural Information Processing Systems}, 32.

\bibitem[Silver and Huffman, 2021]{baseball}
Silver, J. and Huffman, T. (2021).
\newblock Baseball predictions and strategies using explainable {AI}.
\newblock In {\em The 15th Annual MIT Sloan Sports Analytics Conference}.

\bibitem[Simonyan et~al., 2013]{simonyan2013deep}
Simonyan, K., Vedaldi, A., and Zisserman, A. (2013).
\newblock Deep inside convolutional networks: Visualising image classification
  models and saliency maps.
\newblock {\em arXiv preprint arXiv:1312.6034}.

\bibitem[Slack et~al., 2021]{reliable1}
Slack, D., Hilgard, A., Singh, S., and Lakkaraju, H. (2021).
\newblock Reliable post hoc explanations: Modeling uncertainty in
  explainability.
\newblock {\em Advances in Neural Information Processing Systems},
  34:9391--9404.

\bibitem[Smilkov et~al., 2017]{smilkov2017smoothgrad}
Smilkov, D., Thorat, N., Kim, B., Vi{\'e}gas, F., and Wattenberg, M. (2017).
\newblock Smoothgrad: removing noise by adding noise.
\newblock {\em arXiv preprint arXiv:1706.03825}.

\bibitem[Somepalli et~al., 2021]{somepalli2021saint}
Somepalli, G., Goldblum, M., Schwarzschild, A., Bruss, C.~B., and Goldstein, T.
  (2021).
\newblock {SAINT}: Improved neural networks for tabular data via row attention
  and contrastive pre-training.
\newblock {\em arXiv preprint arXiv:2106.01342}.

\bibitem[Springenberg et~al., 2014]{springenberg2014striving}
Springenberg, J.~T., Dosovitskiy, A., Brox, T., and Riedmiller, M. (2014).
\newblock Striving for simplicity: The all convolutional net.
\newblock {\em arXiv preprint arXiv:1412.6806}.

\bibitem[{The White House Office of Science and Technology Policy},
  2022]{aibill}
{The White House Office of Science and Technology Policy} (2022).
\newblock Blueprint for an {AI} bill of rights.
\newblock Accessed: 2022-10-24.

\bibitem[Vansteelandt et~al., 2012]{ps1}
Vansteelandt, S., Bekaert, M., and Claeskens, G. (2012).
\newblock On model selection and model misspecification in causal inference.
\newblock {\em Statistical Methods in Medical Research}, 21(1):7--30.

\bibitem[Wickstr{\o}m et~al., 2020]{clinical}
Wickstr{\o}m, K., Mikalsen, K.~{\O}., Kampffmeyer, M., Revhaug, A., and
  Jenssen, R. (2020).
\newblock Uncertainty-aware deep ensembles for reliable and explainable
  predictions of clinical time series.
\newblock {\em IEEE Journal of Biomedical and Health Informatics},
  25(7):2435--2444.

\bibitem[Wilson et~al., 2019]{predictive1}
Wilson, B., Hoffman, J., and Morgenstern, J. (2019).
\newblock Predictive inequity in object detection.
\newblock {\em arXiv preprint arXiv:1902.11097}.

\bibitem[Xu et~al., 2019]{surv1}
Xu, F., Uszkoreit, H., Du, Y., Fan, W., Zhao, D., and Zhu, J. (2019).
\newblock Explainable {AI}: A brief survey on history, research areas,
  approaches and challenges.
\newblock In {\em CCF International Conference on Natural Language Processing
  and Chinese Computing}, pages 563--574. Springer.

\end{thebibliography}
